\documentclass[journal]{IEEEtran}

\usepackage{booktabs}
\usepackage{array}
\usepackage{listings}
\usepackage{url}
\usepackage{xcolor}
\usepackage{comment}
\usepackage{graphicx}
\usepackage[utf8]{inputenc}
\usepackage{newunicodechar}
\newunicodechar{ɓ}{\textipa{b}}

\begin{document}

\title{Offline AI Modules: Voice-First Offline Architecture, Hardware Reference Stack, and Quantization \& Benchmarking}

\author{
Sunday Afariogun\textsuperscript{1},
Odunolaoluwa Jenrola\textsuperscript{1},
Zeinab Nezami\textsuperscript{2}\\
\textsuperscript{1}Awarri Technologies Limited,
\textsuperscript{2}GSM Association
\thanks{Awarri (https://www.awarri.com/) is an African AI company that builds multilingual language infrastructure for African contexts and voice-first deployment.}
\thanks{The GSM Association (https://www.open-telco.ai/) African AI Language Initiative is an open collaborative programme to develop practical AI tools that serve African language communities.}
}

\maketitle

\begin{abstract}
The Offline AI Modules workstream enables practical, low-power, and
community-accessible deployment of voice-first AI systems that operate
fully offline. Designed for African language communities where speech
is the dominant mode of interaction and internet connectivity is
unreliable or absent, the workstream delivers three reinforcing
components: a modular voice-first offline architecture, a low-cost
hardware reference bill of materials, and a reproducible quantization
and a reproducible quantization and benchmarking pipeline for
instruction-tuned language models in the 2--5\,B parameter class. This paper presents the first end-to-end benchmark
evaluation of the stack across two hardware tiers: an NVIDIA Jetson
Orin NX (Tier~B) and a Raspberry Pi~5 (Tier~A). Three instruction-tuned
models are evaluated across four quantization formats, assessed for
deployment metrics (decode throughput, chat latency, memory, power) and
multilingual quality (topic classification accuracy on MasakhaNEWS
across English, Hausa, Igbo, Nigerian Pidgin, and Yoruba; per-language
perplexity drift). Speech recognition is evaluated using Ethio-ASR on
Amharic and Oromo across both tiers. The principal finding is that
Q4\_K\_M quantization represents the best size-to-quality trade-off for
deployment on both tiers: \texttt{gemma-4-E2B-it} achieves 28.8\,t/s
decode throughput and 89.2\% topic classification accuracy at Q4\_K\_M
on Tier~B, while all three models run within the 16\,GB memory budget
on Tier~A. 
\end{abstract}


\IEEEpeerreviewmaketitle

\section{Introduction}

\IEEEPARstart{T}{he} Offline AI Modules workstream enables practical,
low-power, and community-accessible deployment of voice-first AI systems
that operate fully offline. It exists in direct response to the realities
of African language use and connectivity conditions, where many communities
interact primarily through speech and where internet access is often
limited, unreliable, expensive, or absent.

For many African languages, the dominant mode of interaction is spoken,
not typed. Combined with infrastructure conditions that make always-on
cloud inference impractical, a voice-first offline system is not a
stylistic preference but a functional requirement for inclusion,
usability, and adoption. Offline deployment delivers five concrete
benefits in this context: (i) access in low-connectivity and rural
environments; (ii) resilience under unstable network conditions; (iii) privacy by
keeping voice and language data on-device; (iv) cost reduction by eliminating
per-call cloud inference charges; (v) and local experimentation through
reproducible, low-cost hardware that contributors can build themselves.

The stack combines three reinforcing components: (i) a modular voice-first
offline architecture supporting speech-based interaction end-to-end; (ii) a
low-cost hardware reference stack with a complete bill of materials
across two compute tiers; (iii) and a reproducible quantization and
benchmarking pipeline for small language models in the 2--5\,B parameter
class with shared, comparable metrics. This paper presents the system
design, hardware reference stack, and quantization methodology of the
workstream, together with the first end-to-end benchmark evaluation of
the stack across two hardware tiers: (i) a Raspberry Pi~5 (Tier~A); (ii)
an NVIDIA Jetson Orin NX (Tier~B).

The remainder of this paper is organised as follows.
Section~\ref{sec:arch} describes the voice-first offline system
architecture. Section~\ref{sec:bom} presents the hardware bill of
materials. Section~\ref{sec:quant} details the quantization pipeline
and reference models. Section~\ref{sec:methodology} defines the benchmarking
methodology. Section~\ref{sec:results} reports the evaluation results.
Section~\ref{sec:conclusion} concludes.

\section{Voice-First Offline System Architecture}
\label{sec:arch}

This section describes a practical, deployable offline architecture
supporting voice-first interaction end-to-end on constrained edge
hardware. The design is intentionally modular so that individual
components can be replaced as models and hardware improve, and
sufficiently simple for community contributors to replicate without
specialist tooling.

\subsection{Architecture Overview}

\begin{figure*}[!htbp]
    \centering
    \includegraphics[clip, trim=0cm 0cm 0cm 0.8cm,width=\textwidth]{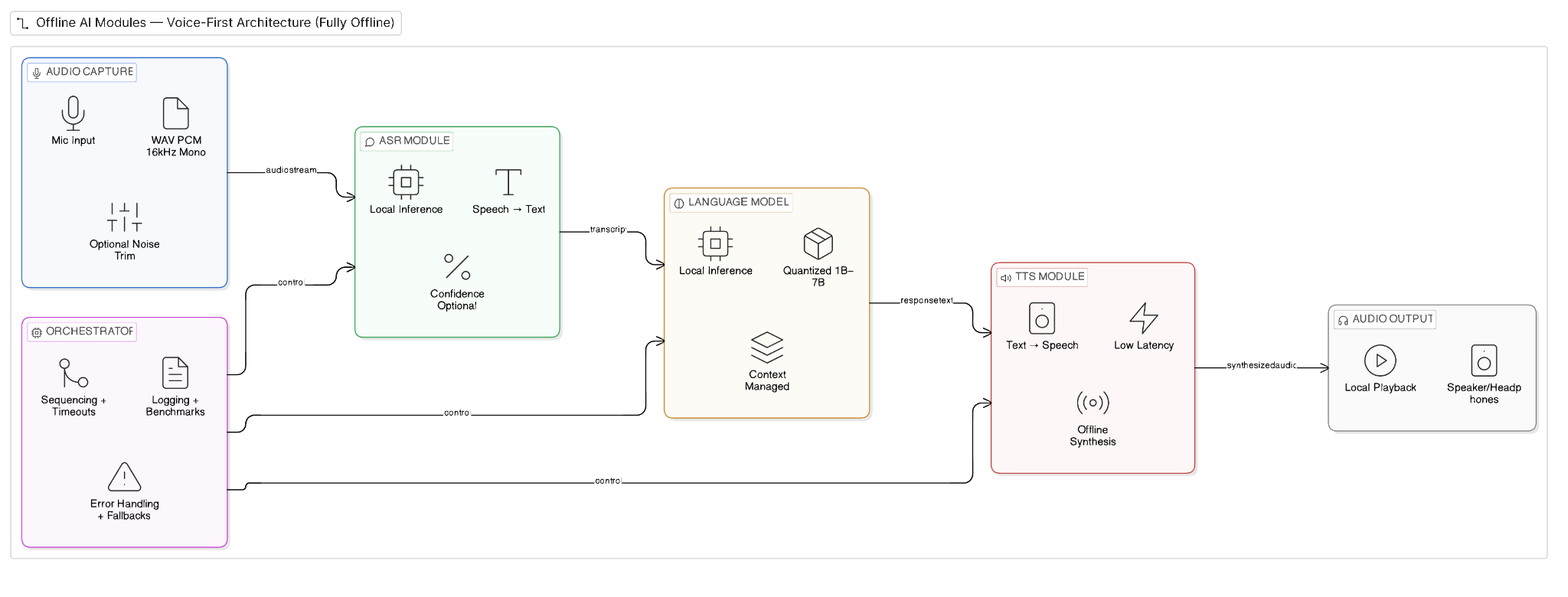}
    \caption{System architecture of the voice-first African language AI framework.}
    \label{fig:arch}
\end{figure*}

The system comprises five independent modules connected through
well-defined interfaces, as illustrated in Fig.~\ref{fig:arch}.
Each module owns a single function and can be substituted without
redesigning the pipeline:

\begin{itemize}
  \item Audio Capture \& Preprocessing
  \item ASR (Speech $\rightarrow$ Text)
  \item Language Model (Text $\rightarrow$ Text)
  \item TTS (Text $\rightarrow$ Speech)
  \item Orchestration \& Control
\end{itemize}

The system operates in two supported configurations:
Configuration~A (CPU-first, low-cost), optimised for 2--5\,B-parameter
models; and Configuration~B (edge-accelerated), optimised for larger
models and lower latency.

\subsection{Module Responsibilities}

\subsubsection{Audio Capture \& Preprocessing}
Captures audio from microphone or file, normalises to a consistent
internal format, and applies optional lightweight preprocessing:
silence trimming, amplitude normalisation, and basic noise suppression.
Preprocessing is intentionally minimal to preserve performance on
CPU-first devices.

\subsubsection{ASR (Automatic Speech Recognition)}
Converts normalised speech to text locally. Returns transcription plus
optional metadata (confidence score, language identifier, segmentation
timestamps). Must run within a predictable time budget on low-cost
devices and support multiple languages without internet connectivity.

\subsubsection{Language Model}
Accepts ASR output text, generates response text locally, and applies
prompt formatting rules to keep behaviour consistent across models.
Must support multiple small-model variants as different communities may
adopt different open-source or locally-built models.

\subsubsection{TTS (Text-to-Speech)}
Converts response text into audio output and returns a WAV waveform
plus playback metadata. Prioritises low latency and intelligibility,
and must be small enough to run locally without GPU acceleration.

\subsubsection{Orchestration \& Control}
Coordinates execution across ASR~$\rightarrow$~LM~$\rightarrow$~TTS,
handles timeouts, fallbacks, and errors, provides consistent logging
for benchmarking and debugging, and exposes a control interface for
contributor testing and automated benchmarking.

\subsection{Interfaces and Data Exchange Standards}

Table~\ref{tab:interfaces} summarises the interface standards used for
data exchange across all system modalities.

\begin{table*}[!t]
\renewcommand{\arraystretch}{1.3}
\caption{Interface Standards}
\label{tab:interfaces}
\centering
\begin{tabular}{lll}
\toprule
\textbf{Data Type} & \textbf{Standard} & \textbf{Notes} \\
\midrule
Audio input  & WAV PCM, 16\,kHz, mono          & All ASR inputs normalised to this format. \\
Audio output & WAV PCM, 16 or 22.05\,kHz, mono & TTS may output 16\,kHz for lightweight playback. \\
Text         & UTF-8                            & All module text payloads must be UTF-8. \\
Metadata     & JSON                             & Timestamps, confidence, language tags, device info. \\
\bottomrule
\end{tabular}
\end{table*}

\subsection{Execution Model}

The system adopts a single-device sequential execution model
(ASR $\rightarrow$ LM $\rightarrow$ TTS) with minimal background
services and benchmarking logging enabled by default. This design
keeps peak memory predictable and simplifies contributor replication.

\subsection{Hardware-Aware Architecture Constraints}

\textit{Tier~A --- Low-Cost CPU-First Devices:} Designed for 2--5\,B-class
language models with context windows limited to 256--512 tokens.
Sequential module execution keeps peak memory predictable. Thermal
and power constraints are acknowledged in runtime behaviour.

\textit{Tier~B --- Edge-Accelerated Devices:} Supports larger models
and lower latency, allows more aggressive batching and longer contexts,
and enables lower-latency interaction loops.

\subsection{Architecture-Level Benchmarking Hooks}

Benchmarking is treated as a first-class architectural feature. The
orchestration layer captures the following on every run:

\begin{itemize}
  \item End-to-end latency (audio in $\rightarrow$ audio out).
  \item Per-module latency (capture, ASR, LM, TTS, playback).
  \item LM throughput (tokens/second) and first-token latency.
  \item Peak RAM usage during the interaction.
  \item Model load time from storage.
\end{itemize}

\section{Hardware Reference Stack}
\label{sec:bom}

This section describes the hardware reference stack used for testing
and benchmarking the offline AI pipeline under realistic deployment
constraints. The stack is documented as a reproducible bill of
materials that community contributors can replicate without specialist
tooling or equipment.

\subsection{Selection Strategy}

Hardware selection follows three criteria: (1)~\textit{reproducibility}: components widely available, no specialised tooling required;
(2)~\textit{coverage}: supports both low-cost CPU-first and
edge-accelerated testing; (3)~\textit{voice-first enablement}: clean
microphone input, speaker output, and a minimal interaction surface.

\subsection{Hardware Bill of Materials}

Table~\ref{tab:bom} summarises the selected hardware components across
the two defined compute tiers.

\begin{table*}[!t]
\renewcommand{\arraystretch}{1.3}
\caption{Hardware Bill of Materials}
\label{tab:bom}
\centering
\begin{tabular}{clll}
\toprule
\textbf{\#} & \textbf{Part} & \textbf{Tier} & \textbf{Role} \\
\midrule
1  & Raspberry Pi 5 (16\,GB RAM)            & A     & Primary CPU-first compute platform. \\
2  & Raspberry Pi AI HAT+                   & A     & Optional NPU acceleration (ASR / small LMs). \\
3  & MicroSD ($\geq$64\,GB, A2 class)       & A     & Boot media; secondary storage for logs. \\
4  & Battery pack (UPS HAT or compatible)   & A     & Stable power for portable / field deployments. \\
5  & Whisplay HAT                           & A     & Integrated mic, speaker, buttons, small display. \\
6  & Waveshare Dual M.2 HAT                 & A     & Adds NVMe storage to Pi~5 for fast model loads. \\
7  & NVMe SSD ($\geq$256\,GB)              & A     & Primary model/dataset storage via Dual M.2 HAT. \\
8  & LLM-8850 M.2 AI Accelerator           & A     & Axera AX8850 SoC, 24\,TOPS @ INT8; offloads \\
   &                                        &       & multimodal LLM and video-analysis workloads. \\
9  & Header spacers and standoffs           & A     & Mechanical assembly for stacked HATs. \\
10 & USB-C 5\,V / 5\,A power supply        & A     & Stable supply for sustained inference workloads. \\
11 & NVIDIA Jetson Orin NX Dev Kit (16\,GB) & B     & Edge-accelerated platform, larger models. \\
12 & NVIDIA Jetson Nano Dev Kit             & B     & Legacy comparison baseline (not required). \\
13 & Active cooling (fan / heatsink)        & A\&B  & Prevents thermal throttling during inference. \\
\bottomrule
\end{tabular}
\end{table*}

\subsection{Functional Mapping}

\subsubsection{CPU-First Low-Cost Platform (Primary Baseline)}
The Raspberry Pi~5 is the baseline environment for running quantised
small models (2--5\,B parameters), validating offline pipeline execution
under strict RAM and thermal limits, and supporting contributor
replication.

\subsubsection{LLM-8850 M.2 AI Accelerator}
The LLM-8850 is an M.2 M-key 2242 card built around the Axera AX8850
SoC, delivering 24\,TOPS at INT8 in a 42\,mm form factor. Mounted on
the Waveshare Dual M.2 HAT, it offloads multimodal and video-analysis
workloads from the Pi~5 CPU, freeing the main cores for the voice-first
pipeline. It also provides a low-cost step up toward Tier~B without
changing the physical platform. Plug-and-play compatibility extends to
RK3588-class SBCs and x86 hosts.

\subsubsection{Edge Acceleration Tier}
The Jetson Orin NX Developer Kit supports higher-performance inference,
faster interaction loops, and benchmarking of larger model sizes.

\subsubsection{Storage and Model Management}
High-speed storage (microSD plus NVMe via the Dual M.2 HAT) ensures
fast model loading, stable throughput, and capacity for multiple
quantised models. Model load time and I/O bottlenecks are common
failure points in offline inference setups.

\subsubsection{Voice Interaction Interface}
The Whisplay HAT bundles microphone, speaker, buttons, and a small
display into a single add-on board, reducing integration overhead and
providing a consistent audio interface across all benchmark runs.

\subsubsection{Power and Mobility}
A battery component supports portability, testing under realistic field
constraints, and validation of power draw under battery operation. Many
target deployment environments lack stable mains electricity.

\subsection{Test Readiness Checks}

Four readiness checks are applied before each benchmark run to ensure
that results reflect model performance rather than hardware instability:

\begin{itemize}
  \item \textbf{Thermal stability} --- sustained inference does not
  cause throttling.
  \item \textbf{Storage performance} --- models load reliably without
  I/O bottlenecks.
  \item \textbf{Audio stability} --- microphone and speaker behave
  consistently across runs.
  \item \textbf{Power stability} --- stable operation on both mains
  and battery.
\end{itemize}

\section{Quantization Pipeline}
\label{sec:quant}

This section describes the quantization workflow used to compress
open-weight and African-built language models into offline-runnable
artifacts for the Tier~A and Tier~B reference platforms. The pipeline
is fully reproducible: each stage is driven by a single shell script
published alongside the workstream documentation.

\subsection{Motivation}

Quantization is the single most important determinant of whether a
given model is deployable on Tier~A hardware. A model in the 3\,B
parameter class occupies approximately 6\,GB of RAM in FP16 and rarely
runs at usable speeds on a Raspberry Pi~5. The same model in Q4\_K\_M
GGUF format occupies approximately 1.8--2.0\,GB and runs at interactive
speeds on CPU. Choosing the right quantization format requires measuring
the trade-off between model quality, file size, memory footprint, and
tokens-per-second throughput on the actual target device.

\subsection{Reference Pipeline}

The reference pipeline takes a Hugging Face checkpoint and produces
GGUF artifacts ready for \texttt{llama.cpp} on both the Raspberry Pi~5
and the Jetson Orin NX. Table~\ref{tab:pipeline} summarises the six
stages of the pipeline; reference commands for the key stages are
provided below.

\begin{table*}[!t]
\renewcommand{\arraystretch}{1.3}
\caption{Quantization Pipeline Stages}
\label{tab:pipeline}
\centering
\begin{tabular}{lll}
\toprule
\textbf{Stage} & \textbf{Tooling} & \textbf{Output} \\
\midrule
1. Acquire checkpoint    & \texttt{huggingface\_hub}           & Local FP16/BF16 safetensors. \\
2. Convert to GGUF (FP16) & \texttt{convert\_hf\_to\_gguf.py}  & \texttt{model-f16.gguf} (intermediate). \\
3. Quantize             & \texttt{llama-quantize}             & Q4\_K\_M, Q5\_K\_M, Q8\_0 GGUFs. \\
4. Smoke-test           & \texttt{llama-cli}, fixed prompt set & Pass/fail per quantization. \\
5. Benchmark            & \texttt{llama-bench} + harness      & JSON + CSV per schema in Section~\ref{sec:methodology}. \\
6. Publish              & HF Hub or shared object store       & Versioned GGUFs with model card. \\
\bottomrule
\end{tabular}
\end{table*}

Stage~2 --- convert FP16 safetensors to GGUF:
\begin{lstlisting}[language=bash]
python convert_hf_to_gguf.py ./model-hf \
  --outfile ./model-f16.gguf \
  --outtype f16
\end{lstlisting}

Stage~3 --- produce three quantization variants:
\begin{lstlisting}[language=bash]
for QT in Q4_K_M Q5_K_M Q8_0; do
  ./llama-quantize ./model-f16.gguf \
    ./model-${QT}.gguf ${QT}
done
\end{lstlisting}

Stage~5 --- run the standard benchmark evaluation on the target device:
\begin{lstlisting}[language=bash]
./llama-bench -m ./model-Q4_K_M.gguf \
  -p 32,64 -n 64,128 -t 4 -ngl 0 \
  -o json > bench-Q4_K_M.json
\end{lstlisting}

\subsection{Supported Quantization Formats}
Three GGUF k-quant formats are supported as the primary evaluation
targets, spanning the realistic range a contributor would consider on
Tier~A hardware. Table~\ref{tab:quant} summarises their characteristics.
Q4\_K\_M is the recommended default for community deployments because
it represents the best size-to-quality ratio currently available for
CPU inference.

\begin{table*}[!t]
\renewcommand{\arraystretch}{1.3}
\caption{Supported GGUF Quantization Formats}
\label{tab:quant}
\centering
\begin{tabular}{lllll}
\toprule
\textbf{Format} & \textbf{Bits/wt (avg)} & \textbf{Size (3B)} & \textbf{Quality vs FP16} & \textbf{Use} \\
\midrule
Q4\_K\_M & $\sim$4.85 & $\sim$1.9\,GB & Task accuracy preserved (see Sec.~\ref{sec:results}) & Default, Pi~5 (Tier~A). \\
Q5\_K\_M & $\sim$5.7  & $\sim$2.2\,GB & Near-FP16 for most prompts       & Pi~5 + NVMe; better quality. \\
Q8\_0    & $\sim$8.5  & $\sim$3.3\,GB & Effectively lossless             & Jetson Orin NX (Tier~B). \\
F16      & 16         & $\sim$6.0\,GB & Reference                        & Tier~B; calibration baseline. \\
\bottomrule
\end{tabular}
\end{table*}

Q4\_K\_M is the recommended default for community deployments because it represents the best size-to-quality ratio currently available for CPU inference.

\subsection{Quality Validation}

Each quantised artifact is validated along three axes before inclusion
in the evaluation:

\begin{itemize}
  \item \textbf{Perplexity drift} --- measured against held-out text
  for the full-precision reference and each quantization, reported per
  language. Drift is treated as a diagnostic to be read alongside
  task-level metrics rather than a pass/fail gate: the evaluation in
  Section~\ref{sec:results} shows drift of up to 6.4\% at Q4\_K\_M
  coexisting with stable task accuracy. Q8\_0 drift is expected to be
  near-zero (under 0.8\% observed across all models and languages).

  \item \textbf{Smoke-test pass rate} --- a fixed set of 20 prompts
  covering greetings, intent classification, simple QA, and short
  instructions in each supported language, marked pass / partial / fail.

  \item \textbf{Behavioural diff} --- quantised outputs compared to the
  full-precision reference on the same prompt set. High divergence at
  Q4\_K\_M is a signal to evaluate Q5\_K\_M as a fallback for that
  model.
\end{itemize}

\subsection{Reference Models}

Table~\ref{tab:models} lists the three open-weight models selected as
the primary evaluation targets. Selection criteria included
instruction-tuning, parameter count compatible with Tier~A deployment,
permissive licensing, and multilingual pre-training coverage relevant
to African languages.

\begin{table*}[!t]
\renewcommand{\arraystretch}{1.3}
\caption{Reference Models for Quantization and Evaluation}
\label{tab:models}
\centering
\begin{tabular}{llll}
\toprule
\textbf{Model} & \textbf{Params} & \textbf{License} & \textbf{Why included} \\
\midrule
gemma-4-E2B-it      & 5\,B total / 2\,B eff. & Gemma Terms of Use &
Mixture-of-experts architecture; \\
                    &                        &                    &
highest accuracy in evaluation. \\
gemma-2-2b-it       & 2.6\,B                 & Gemma Terms of Use &
Smallest memory footprint; \\
                    &                        &                    &
viable on memory-constrained Tier~A. \\
Qwen2.5-3B-Instruct & 3.1\,B                 & Apache 2.0         &
Permissive license; strong multilingual \\
                    &                        &                    &
coverage; lowest PPL drift at Q4\_K\_M. \\
\bottomrule
\end{tabular}
\end{table*}

As African-built models are released by parallel workstreams, they will
be added using the same pipeline, evaluation prompts, and publication
format.

\section{Benchmarking Methodology}
\label{sec:methodology}

This section defines the metrics, harness, and reporting format used
to evaluate the offline AI stack. A shared methodology ensures that
results across devices, models, and contributors are directly
comparable and can be aggregated into a consistent picture of
deployment feasibility. Table~\ref{tab:metrics} summarises the
benchmark metrics used to evaluate system performance across latency,
throughput, and resource usage. Table~\ref{tab:conditions} defines the
standard test conditions applied to ensure reproducibility.

\begin{table*}[!t]
\renewcommand{\arraystretch}{1.3}
\caption{Benchmark Metrics}
\label{tab:metrics}
\centering
\begin{tabular}{lll}
\toprule
\textbf{Metric} & \textbf{Unit} & \textbf{Definition} \\
\midrule
End-to-end latency     & ms         & Audio-in start to audio-out start. \\
ASR latency            & ms         & End-of-speech to ASR text output. \\
LM first-token latency & ms         & Prompt submission to first generated token. \\
LM throughput          & tokens/s   & Sustained rate over $\geq$64 generated tokens. \\
TTS latency            & ms         & Text input to first audio sample available. \\
Peak RAM               & MB         & Maximum resident memory during interaction. \\
Model load time        & s          & Process start to model ready (cold cache). \\
\bottomrule
\end{tabular}
\end{table*}


\begin{table*}[!t]
\renewcommand{\arraystretch}{1.3}
\caption{Standard Test Conditions}
\label{tab:conditions}
\centering
\begin{tabular}{lll}
\toprule
\textbf{Condition} & \textbf{Default} & \textbf{Rationale} \\
\midrule
Network          & Disconnected                        & Enforces the offline guarantee. \\
Background load  & Idle desktop, no other workloads    & Removes contention from the measurement. \\
Storage state    & Cold cache for load; warm for runs  & Separates load-time from inference effects. \\
Thermal state    & 5-min idle before run; ambient $\leq$28°C & Prevents thermal bias. \\
Power            & Documented (mains vs battery)       & Battery can throttle or affect clocks. \\
Run length       & $\geq$10 consecutive interactions   & Smooths out one-off outliers. \\
Warm-up          & Discard first interaction           & Load time reported separately. \\
\bottomrule
\end{tabular}
\end{table*}

\subsection{Logging Schema (JSON)}

Every run produces one run-header record (Fig.~\ref{lst:run_header}) and one metrics record per interaction (Fig.~\ref{lst:interaction_metrics}).
\begin{figure*}[!t]
\begin{lstlisting}[breaklines=true, columns=fullflexible, basicstyle=\ttfamily\footnotesize]
{
    "run_id": "2026-07-21T22:34:17Z_pi5_e2b_q4km",
    "device": { "name": "Raspberry Pi 5", "ram_gb": 16, "storage": "NVMe",
                "os": "Raspberry Pi OS 12 (bookworm)", "tier": "A", "cooling": "active fan" },
    "pipeline": { "mode": "voice_to_voice", "asr_engine": "whisper.cpp 1.6.x",
                  "lm_runtime": "llama.cpp build 9670", "tts_engine": "piper 1.2.x" },
    "model": { "name": "gemma-4-E2B-it", "params_b": 5.0, "format": "GGUF",
               "quantization": "Q4_K_M", "context_tokens": 512 },
    "timestamp_utc": "2026-07-21T22:34:17Z"
}
\end{lstlisting}
\caption{Run header JSON record.}
\label{lst:run_header}
\end{figure*}

\begin{figure*}[!t]
\begin{lstlisting}[breaklines=true, columns=fullflexible, basicstyle=\ttfamily\footnotesize]
{
    "run_id": "2026-07-21T22:34:17Z_pi5_e2b_q4km",
    "interaction_id": "0001",
    "input": { "audio_duration_ms": 3400, "language": "auto" },
    "metrics": {
      "end_to_end_latency_ms": 8970,
      "asr": { "latency_ms": 2100, "confidence": 0.86 },
      "lm": { "prompt_tokens": 62, "tokens_generated": 128,
              "first_token_latency_ms": 2019, "tokens_per_second": 7.5, "peak_ram_mb": 2600 },
      "tts": { "latency_ms": 1450, "output_audio_duration_ms": 4200 }
    },
    "errors": [],
    "timestamp_utc": "2026-07-21T22:34:41Z"
}
\end{lstlisting}
\caption{Per-interaction metrics JSON record.}
\label{lst:interaction_metrics}
\end{figure*}

\subsection{CSV Logging Schema}

A flat CSV is provided for contributors who prefer spreadsheets. One row per interaction; the header is fixed:

\begin{lstlisting}
run_id, interaction_id, device, ram_gb, model_name,
params_b, quant, context_tokens, asr_engine,
tts_engine, end_to_end_latency_ms, asr_latency_ms,
lm_first_token_latency_ms, lm_tokens_generated,
lm_tokens_per_second, tts_latency_ms, peak_ram_mb,
model_load_time_s, errors
\end{lstlisting}

\subsection{Statistical Protocol}
The following statistical procedures are applied to quality evaluation
results reported in Section~\ref{sec:results}.   Topic classification accuracy is reported with 95\% Wilson score   intervals (approximately $\pm 3.5$ percentage points at 85\% accuracy on $n = 400$ items). To test whether quantization changes task accuracy, each quantized format is compared against its own full-precision reference using an exact McNemar test on paired per-item outcomes, with Holm correction applied across the nine planned comparisons (three models $\times$ three quantization formats). A non-significant result indicates that the evaluation cannot resolve a difference at this sample size; it does not establish equivalence.

\section{Benchmark Results}
\label{sec:results}

This section reports the benchmark evaluation for the Offline AI Modules
stack. Two device tiers were evaluated under the standard test conditions
defined in Table~\ref{tab:conditions}: Tier~B (an NVIDIA Jetson Orin NX
16\,GB, Ampere GPU) and Tier~A (a Raspberry Pi~5 16\,GB, quad-core
Cortex-A76, CPU-only). The Tier~B evaluation covers three instruction-tuned
models, \texttt{gemma-4-E2B-it} (5\,B total / 2\,B effective parameters),
\texttt{gemma-2-2b-it} (2.6\,B), and \texttt{Qwen2.5-3B-Instruct} (3.1\,B),
each assessed across four precision levels: a full-precision reference
(BF16 for \texttt{gemma-4-E2B-it}, F16 for the others), Q8\_0, Q5\_K\_M,
and Q4\_K\_M. Deployment metrics reported include decode throughput,
prefill throughput, chat time-to-first-token (TTFT), warm-cache server
startup time, process resident memory (RSS), and peak power draw.
Quality is assessed through two complementary measures: topic
classification accuracy on the MasakhaNEWS benchmark~\cite{masakhanews}
(400 stratified items across English, Hausa, Igbo, Nigerian Pidgin, and
Yoruba; chance level 25\%) and per-language perplexity drift relative to
the full-precision reference. Tier~A deployment metrics were collected
under an earlier harness revision and are reported for feasibility
context; they were not re-measured under the revised protocol applied
to Tier~B.

\begin{figure}[!t]
  \centering
  \includegraphics[width=\columnwidth]{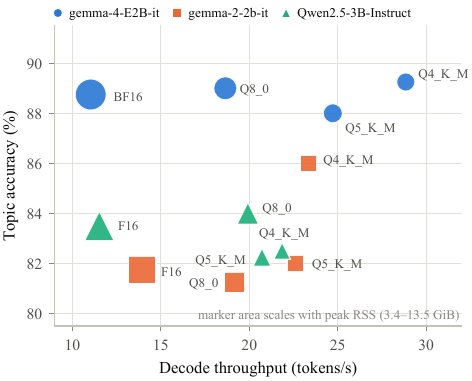}
  \caption{Quality--cost trade-off on Tier~B: topic accuracy versus
  decode throughput, with marker area proportional to peak process RSS
  (3.4--13.5\,GiB). \texttt{gemma-4-E2B-it} Q4\_K\_M dominates the
  upper-right frontier.}
  \label{fig:quality_cost}
\end{figure}

The principal finding is that Q4\_K\_M represents the best
size-to-quality trade-off for deployment on both tiers, as illustrated
in Fig.~\ref{fig:quality_cost}. \texttt{gemma-4-E2B-it} achieves the
highest decode throughput on Tier~B at 28.8\,t/s and the highest topic
classification accuracy at 89.2\%, with a macro PPL drift of only 0.63\%
relative to its BF16 reference. Its relative ranking among the three
models is consistent across deployment metrics and quality measures,
confirming that its efficiency and accuracy advantages are not
format-dependent artefacts.

\subsection{Benchmark Datasets}

The quality evaluation uses five held-out corpora drawn from the
MasakhaNEWS dataset~\cite{masakhanews}, one per language: English (\texttt{eng}),
Hausa (\texttt{hau}), Igbo (\texttt{ibo}), Nigerian Pidgin (\texttt{pcm}),
and Yoruba (\texttt{yor}). Each corpus contains approximately 50,000
whitespace-delimited words, providing a consistent scale for perplexity
measurement across languages. Article count and length vary by language,
reflecting differences in news article conventions: English comprises
95 longer articles (mean 531 words), while Hausa, Igbo, Nigerian Pidgin,
and Yoruba each contain 109--154 shorter articles (mean 325--461 words).
The topic classification evaluation uses a stratified subset of 400 items,
80 per language, balanced across four classes (entertainment, health,
politics, sports), giving a chance-level accuracy of 25\%.

The corpora differ meaningfully in orthographic complexity, which is
relevant to tokenization efficiency and perplexity interpretation.
English uses only 9 non-ASCII characters; Hausa and Igbo use
22 and 20 respectively, including tone-marked vowels and implosive
consonant characters specific to their phonological inventories;
Yoruba uses 40 unique non-ASCII characters, the highest among the
five languages, reflecting its extensive use of diacritically marked
vowels and nasal tones. This orthographic variation means that raw
perplexity values are not directly comparable across language families;
results are therefore reported as drift relative to each model's
full-precision reference, isolating the effect of quantization from
baseline tokenization differences.

\subsection{Tier B (Jetson Orin NX, GPU): Full Quantization Evaluation}

Table~\ref{tab:tierB} presents the complete Tier~B evaluation results
across all model and quantization combinations. Deployment metrics and
quality measures are reported jointly to enable format selection that
accounts for both inference cost and task performance.
Decode and prefill throughput are measured via the llama.cpp server. TTFT is the
end-to-end chat time-to-first-token (median) from HTTP request after
one warm-up. Startup reports warm-cache server readiness, not cold
disk load. RSS is the process resident memory, which is the preferred
measure on the Jetson's unified-memory architecture. Power reports
VDD\_IN peak. Quality is assessed via MasakhaNEWS four-class topic
classification over 400 stratified items (chance = 25\%); 95\% Wilson
confidence intervals are omitted here for space. The reference precision used for PPL
drift is BF16 for \texttt{gemma-4-E2B-it} and F16 for the other two
models. Bold rows indicate the recommended deployment format per model.

\begin{figure}[!t]
  \centering
  \includegraphics[width=\columnwidth]{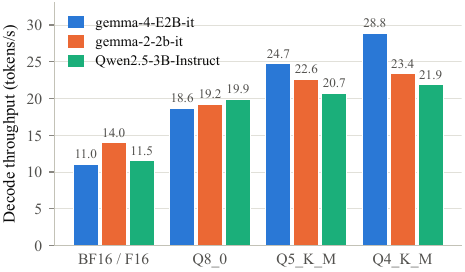}
  \caption{Decode throughput on Tier~B (Jetson Orin NX, GPU) by precision
  level. Quantization to Q4\_K\_M yields $1.7$--$2.6\times$ the
  full-precision decode rate.}
  \label{fig:decode}
  \end{figure}
  
\subsubsection{Deployment Metrics}

Q4\_K\_M consistently achieves the highest decode throughput and smallest
  memory footprint across all three models (Fig.~\ref{fig:decode}). Transitioning from Q4\_K\_M to
Q8\_0 incurs approximately 35\% decode throughput degradation for
\texttt{gemma-4-E2B-it} (28.8~$\rightarrow$~18.6\,t/s) and approximately
18\% for \texttt{gemma-2-2b-it} (23.4~$\rightarrow$~19.2\,t/s).
Full-precision references are not viable for on-device deployment:
\texttt{gemma-2-2b-it} F16 achieves only 14.0\,t/s with 10,237\,MiB
resident memory, and \texttt{gemma-4-E2B-it} BF16 requires 13,508\,MiB
at 11.0\,t/s. Chat TTFT for the deployable quantization formats
(Q4\_K\_M, Q5\_K\_M, Q8\_0) ranges from 446 to 501\,ms at the median,
remaining within an acceptable threshold for interactive voice agents.
Peak power consumption held stable at 20--24\,W across all configurations,
with no thermal throttling observed.

\subsubsection{Quality Evaluation}

\begin{figure}[!t]
  \centering
  \includegraphics[width=\columnwidth]{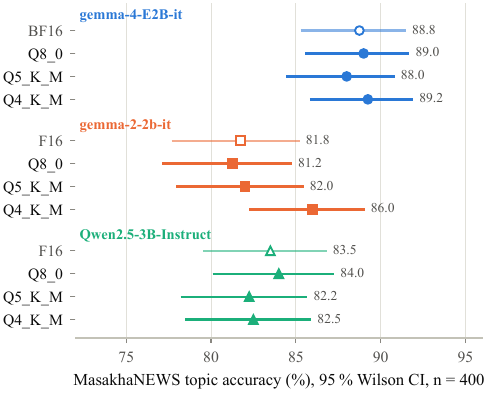}
  \caption{MasakhaNEWS topic accuracy with 95\% Wilson intervals ($n=400$)
  for all model--precision combinations on Tier~B. Hollow markers denote
  full-precision references. \texttt{gemma-2-2b-it} Q4\_K\_M is the only
  quantization differing significantly from its reference (exact McNemar,
  Holm-corrected $p=0.030$) --- an improvement.}
  \label{fig:acc_forest}
\end{figure}

\begin{figure}[!t]
  \centering
  \includegraphics[width=\columnwidth]{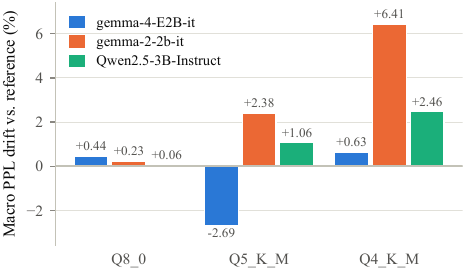}
  \caption{Macro perplexity drift of each quantization relative to its own
  full-precision reference. Drift does not track task accuracy:
  \texttt{gemma-2-2b-it} Q4\_K\_M shows the largest drift ($+6.41\%$) yet
  the largest accuracy gain (Fig.~\ref{fig:acc_forest}).}
  \label{fig:ppl_drift}
\end{figure}

Topic classification accuracy on MasakhaNEWS was evaluated across five
  African languages; Fig.~\ref{fig:acc_forest} shows all twelve
  model--precision combinations with their 95\% Wilson intervals. \texttt{gemma-4-E2B-it} achieves the highest overall
accuracy at 89.2\% (Q4\_K\_M), with macro F1 of 89.3\% and a PPL drift
of only 0.63\% relative to BF16, confirming that quantization introduces
negligible quality degradation for this model. \texttt{gemma-2-2b-it}
  Q4\_K\_M achieves 86.0\% accuracy with a macro F1 of 86.4\%. Notably,
  this is the only configuration in the study whose accuracy differs
  significantly from its full-precision reference (exact McNemar
  $p = 0.0033$, Holm-corrected $p = 0.030$), and the difference is an
  improvement of $+4.25$ percentage points over F16 (81.8\%). Its PPL
  drift of 6.41\% is simultaneously the highest among the deployable
    formats, illustrating that perplexity drift neither predicts nor
    bounds task-level accuracy change (Fig.~\ref{fig:ppl_drift}). No other quant--reference pair
  differs significantly after correction.
\texttt{Qwen2.5-3B-Instruct} Q4\_K\_M achieves 82.5\% accuracy with a
PPL drift of 2.46\%, and its accuracy is stable across all quantization levels (82.2--84.0\%).

\begin{table*}[!t]
\renewcommand{\arraystretch}{1.3}
\caption{Tier B (Jetson Orin NX) LLM Benchmark --- Deployment Metrics and Quality}
\label{tab:tierB}
\centering
\begin{tabular}{llrrrrrrrrr}
\toprule
\textbf{Model} & \textbf{Quant} & \textbf{Size} & \textbf{Decode} & \textbf{Prefill} & \textbf{TTFT} & \textbf{Startup} & \textbf{RSS} & \textbf{Power} & \textbf{Accuracy} & \textbf{Macro F1} \\
 & & \textbf{(GiB)} & \textbf{(t/s)} & \textbf{(t/s)} & \textbf{med (ms)} & \textbf{(s)} & \textbf{(MiB)} & \textbf{(W)} & \textbf{(\%)} & \textbf{(\%)} \\
\midrule
Gemma-4-E2B-it & BF16    & 8.67 & 11.0 & 601.1 & 550 & 11.07 & 13508 & 20.9 & 88.8 & 88.9 \\
Gemma-4-E2B-it          & Q8\_0   & 4.70 & 18.6 & 972.0 & 466 &  5.03 &  7565 & 20.3 & 89.0 & 89.2 \\
Gemma-4-E2B-it          & Q5\_K\_M & 3.13 & 24.7 & 909.7 & 487 &  4.52 &  5177 & 21.2 & 88.0 & 88.2 \\
\textbf{Gemma-4-E2B-it} & \textbf{Q4\_K\_M} & \textbf{2.89} & \textbf{28.8} & 940.5 & \textbf{477} & 4.52 & \textbf{4701} & 21.2 & \textbf{89.2} & \textbf{89.3} \\
\midrule
Gemma-2-2b-it  & F16     & 4.88 & 14.0 & 896.6 & 425 &  5.03 & 10237 & 23.4 & 81.8 & 82.3 \\
Gemma-2-2b-it           & Q8\_0   & 2.59 & 19.2 &1033.1 & 446 &  3.52 &  5563 & 21.8 & 81.2 & 81.8 \\
Gemma-2-2b-it           & Q5\_K\_M & 1.79 & 22.6 & 937.2 & 501 &  3.27 &  3924 & 23.1 & 82.0 & 82.7 \\
\textbf{Gemma-2-2b-it}  & \textbf{Q4\_K\_M} & \textbf{1.59} & \textbf{23.4} & 961.6 & 491 & 3.09 & \textbf{3515} & 22.5 & \textbf{86.0} & \textbf{86.4} \\
\midrule
Qwen2.5-3B     & F16     & 5.75 & 11.5 & 746.7 & 497 &  4.60 & 12040 & 23.0 & 83.5 & 84.0 \\
Qwen2.5-3B              & Q8\_0   & 3.06 & 19.9 & 872.7 & 456 &  3.27 &  6522 & 23.8 & 84.0 & 84.6 \\
Qwen2.5-3B              & Q5\_K\_M & 2.07 & 20.7 & 749.8 & 499 &  2.77 &  4504 & 23.4 & 82.2 & 82.6 \\
\textbf{Qwen2.5-3B}     & \textbf{Q4\_K\_M} & \textbf{1.80} & \textbf{21.9} & 771.2 & 487 & 2.77 & \textbf{3936} & 23.2 & 82.5 & 82.9 \\
\bottomrule
\end{tabular}
\end{table*}

Table~\ref{tab:lang_acc_ppl} reports per-language topic classification
accuracy and perplexity drift for all evaluated model and quantization
combinations. Accuracy is largely stable across quantization levels for
\texttt{gemma-4-E2B-it}, with Nigerian Pidgin consistently achieving the
highest per-language score (95.0--96.2\%) across all formats, reflecting
strong coverage of West African English-influenced language patterns.
Hausa is the weakest language for \texttt{gemma-4-E2B-it} and
  \texttt{Qwen2.5-3B-Instruct}, while \texttt{gemma-2-2b-it} is weakest
  on English; the Hausa deficit is consistent with more limited 
  representation in pre-training corpora. PPL drift at Q4\_K\_M is small and uniformly positive for \texttt{Qwen2.5-3B-Instruct} (+1.35 to +4.20\%), mixed in sign and smallest in aggregate for \texttt{gemma-4-E2B-it} (-4.27 to +5.97\%; macro +0.63\%), and highest for \texttt{gemma-2-2b-it} (+3.27 to +8.29\%), particularly on Hausa and Igbo. Negative drift values
observed for \texttt{gemma-4-E2B-it} at Q5\_K\_M are within measurement
uncertainty and should not be interpreted as genuine quality improvement
over the BF16 reference.

\begin{table*}[!t]
\renewcommand{\arraystretch}{1.2}
\caption{Tier B (Jetson Orin NX) --- Topic Accuracy (\%) and PPL Drift (\%) by Language}
\label{tab:lang_acc_ppl}
\centering
\begin{tabular}{llrrrrrrrrrr}
\toprule
 & & \multicolumn{5}{c}{\textbf{Accuracy (\%)}} & \multicolumn{5}{c}{\textbf{PPL Drift (\%)}} \\
\cmidrule(lr){3-7}\cmidrule(lr){8-12}
\textbf{Model} & \textbf{Quant} & \textbf{eng} & \textbf{hau} & \textbf{ibo} & \textbf{pcm} & \textbf{yor} & \textbf{eng} & \textbf{hau} & \textbf{ibo} & \textbf{pcm} & \textbf{yor} \\
\midrule
Gemma-4-E2B-it & BF16    & 88.8 & 83.8 & 88.8 & 95.0 & 87.5 & 0.00  &  0.00 &  0.00 &  0.00 &  0.00 \\
Gemma-4-E2B-it & Q8\_0   & 88.8 & 83.8 & 88.8 & 95.0 & 88.8 & +0.08 & +0.68 & +0.69 & -0.05 & +0.77 \\
Gemma-4-E2B-it & Q5\_K\_M & 86.2 & 83.8 & 86.2 & 96.2 & 87.5 & -5.21 & -2.33 & -1.01 & -3.37 & -1.52 \\
Gemma-4-E2B-it & Q4\_K\_M & 88.8 & 85.0 & 87.5 & 96.2 & 88.8 & +2.77 & -0.65 & -0.67 & +5.97 & -4.27 \\
\midrule
Gemma-2-2b-it  & F16     & 77.5 & 78.8 & 83.8 & 82.5 & 86.2 & 0.00  &  0.00 &  0.00 &  0.00 &  0.00 \\
Gemma-2-2b-it  & Q8\_0   & 77.5 & 78.8 & 82.5 & 81.2 & 86.2 & +0.24 & +0.57 & +0.23 & +0.13 & +0.01 \\
Gemma-2-2b-it  & Q5\_K\_M & 78.8 & 78.8 & 81.2 & 83.8 & 87.5 & +0.45 & +3.43 & +3.04 & +2.50 & +2.46 \\
Gemma-2-2b-it  & Q4\_K\_M & 82.5 & 83.8 & 88.8 & 88.8 & 86.2 & +3.27 & +8.29 & +8.26 & +5.88 & +6.37 \\
\midrule
Qwen2.5-3B     & F16     & 86.2 & 73.8 & 86.2 & 87.5 & 83.8 & 0.00  &  0.00 &  0.00 &  0.00 &  0.00 \\
Qwen2.5-3B     & Q8\_0   & 86.2 & 76.2 & 86.2 & 87.5 & 83.8 & +0.12 & +0.08 & -0.19 & +0.08 & +0.19 \\
Qwen2.5-3B     & Q5\_K\_M & 85.0 & 73.8 & 83.8 & 88.8 & 80.0 & +0.83 & +1.60 & +0.78 & +0.57 & +1.51 \\
Qwen2.5-3B     & Q4\_K\_M & 86.2 & 76.2 & 83.8 & 90.0 & 76.2 & +1.97 & +3.20 & +1.35 & +1.59 & +4.20 \\
\bottomrule
\end{tabular}
\begin{minipage}{\textwidth}
\vspace{4pt}
\footnotesize\textit{Accuracy: correct/80 per language (chance = 25\%). PPL drift: percentage
change relative to the full-precision reference (BF16 for \texttt{gemma-4-E2B-it},
F16 for others); negative values are within measurement uncertainty and do not
indicate genuine improvement. Languages: eng = English, hau = Hausa,
ibo = Igbo, pcm = Nigerian Pidgin, yor = Yoruba.}
\end{minipage}
\end{table*}

Per-class precision, recall, and F1 are reported in Appendix~\ref{app:class_f1}.

\subsection{Tier A (Raspberry Pi 5, CPU): Deployable Q4\_K\_M Evaluation}

Table~\ref{tab:tierA} presents Q4\_K\_M results on the Raspberry Pi~5. All three models operate within the 16\,GB memory budget (peak resident memory ranges from 2.3 to 2.6\,GB), marginally lower than on the Jetson Orin NX owing to the absence of GPU-mirrored memory allocation. The relative throughput ranking among models is preserved from Tier~B: \texttt{gemma-4-E2B-it} achieves the highest decode rate at 7.5\,t/s, followed by \texttt{gemma-2-2b-it} at 6.8\,t/s and \texttt{Qwen2.5-3B-Instruct} at 6.1\,t/s. All smoke-test outputs were well-formed, and no thermal throttling was observed during the evaluation.

As Tier~A uses the identical Q4\_K\_M GGUF artifacts evaluated on
Tier~B, quality figures (accuracy, macro F1, and per-language PPL
drift) are those reported for Q4\_K\_M in Tables~\ref{tab:tierB}
and~\ref{tab:lang_acc_ppl} and are not repeated here.

\begin{table*}[!t]
\renewcommand{\arraystretch}{1.3}
\caption{Tier A (Raspberry Pi 5, CPU) LLM Benchmark, Q4\_K\_M}
\label{tab:tierA}
\centering
\begin{tabular}{lrrrr}
\toprule
\textbf{Model} & \textbf{Decode (t/s)} & \textbf{Prefill (t/s)} & \textbf{First-response (ms)} & \textbf{Peak RAM (MB)} \\
\midrule
Gemma-4-E2B-it      & 7.5 & 31.7 & 2019 & 2278 \\
Gemma-2-2b-it       & 6.8 & 39.4 & 1624 & 2428 \\
Qwen2.5-3B-Instruct & 6.1 & 29.4 & 2180 & 2578 \\
\bottomrule
\end{tabular}
\end{table*}

\subsection{Cross-Tier Comparison (Q4\_K\_M)}

Table~\ref{tab:crossdevice} presents a direct comparison of the Q4\_K\_M
quantization format across both hardware tiers. The performance penalty
incurred by CPU-only inference is highly asymmetric across operation
types. Decode throughput, which is memory-bandwidth-bound, degrades by a
factor of 3.4--3.8$\times$, yielding a generation rate of 6.1--7.5\,t/s
on Tier~A, which remains usable for streaming voice output. By contrast,
prefill throughput, which is compute-bound and relies on parallel matrix
operations, degrades by a factor of 24--30$\times$ on four ARM
Cortex-A76 cores. This asymmetry concentrates the performance impact in
response initiation latency: on Tier~A, first-response latency rises to
1.6--2.2\,s, exceeding the perceptual threshold for interactive voice
response. On Tier~B, chat TTFT ranges from 477 to 491\,ms at the median
for Q4\_K\_M across the three models; this metric is not directly
comparable to Tier~A first-response latency as it includes the full HTTP
chat path, but both point to the same bottleneck: prefill on
compute-constrained hardware. The primary optimisation target is
therefore prefill reduction: minimising system prompt length, constraining
retrieval context, and employing prompt caching where applicable.

\begin{table*}[!t]
\renewcommand{\arraystretch}{1.3}
\caption{Jetson Orin NX (GPU) vs.\ Raspberry Pi 5 (CPU), Q4\_K\_M, Per Model}
\label{tab:crossdevice}
\centering
\begin{tabular}{llrrrp{4.2cm}}
\toprule
\textbf{Model} & \textbf{Metric} & \textbf{Jetson (GPU)} & \textbf{Pi 5 (CPU)} & \textbf{Ratio} & \textbf{Note} \\
\midrule
Gemma-4-E2B & Decode (t/s)    & 28.8 & 7.5  & 3.8$\times$  & \\
Gemma-4-E2B & Prefill (t/s)   & 940.5 & 31.7 & 29.7$\times$ & \\
Gemma-4-E2B & Chat TTFT (ms)  & 477  & ---  & ---          & Tier~A uses raw pipeline; not directly comparable \\
\midrule
Gemma-2-2b  & Decode (t/s)    & 23.4 & 6.8  & 3.4$\times$  & \\
Gemma-2-2b  & Prefill (t/s)   & 961.6 & 39.4 & 24.4$\times$ & \\
Gemma-2-2b  & Chat TTFT (ms)  & 491  & ---  & ---          & Tier~A uses raw pipeline; not directly comparable \\
\midrule
Qwen2.5-3B  & Decode (t/s)    & 21.9 & 6.1  & 3.6$\times$  & \\
Qwen2.5-3B  & Prefill (t/s)   & 771.2 & 29.4 & 26.2$\times$ & \\
Qwen2.5-3B  & Chat TTFT (ms)  & 487  & ---  & ---          & Tier~A uses raw pipeline; not directly comparable \\
\bottomrule
\end{tabular}
\begin{minipage}{\textwidth}
\vspace{4pt}
\footnotesize\textit{Tier~B decode and prefill from llama.cpp server benchmark. Chat TTFT
is the median end-to-end HTTP request latency after one warm-up, available
for Tier~B only. Tier~A first-response latency (1.6--2.2\,s) is measured
via the raw pipeline harness and is reported separately in
Table~\ref{tab:tierA}. Ratios are computed on the overlapping metrics only.}
\end{minipage}
\end{table*}

\subsection{Speech Recognition (Ethio-ASR): Both Tiers}

The ASR evaluation assessed two Ethio-ASR multilingual
models~\cite{ethioasr} (94\,M and 600\,M parameters) on the FLEURS Amharic and Oromo test sets. Recognition quality is largely device-independent: normalised word error rate (WER\textsubscript{norm}) on Amharic varies by only 0.288~$\rightarrow$~0.291 for the 94\,M model and 0.228~$\rightarrow$~0.252 for the 600\,M model between GPU and CPU inference, differences attributable to fp16~$\rightarrow$~fp32 numeric variation and limited sample size rather than a systematic regression.
\begin{figure*}[!htbp]
    \centering
    \includegraphics[clip, trim=0.5cm 11.5cm 3cm 4.5cm,width=\textwidth]{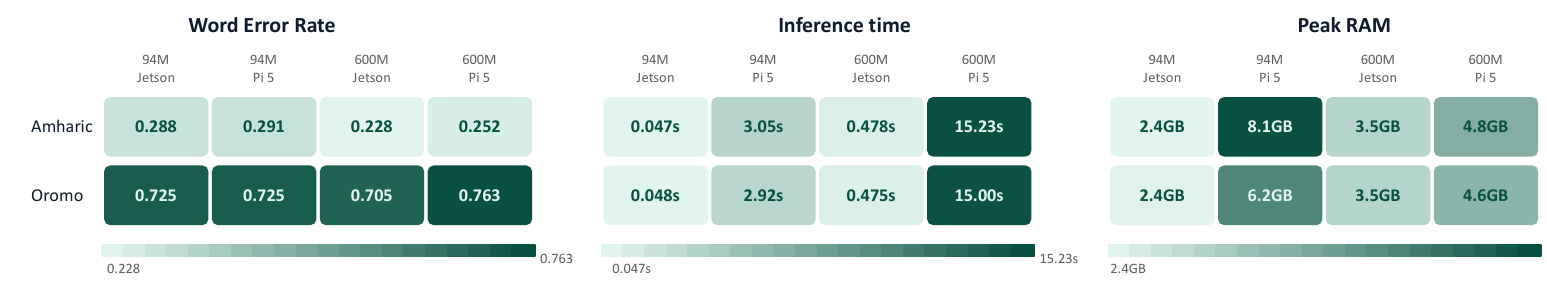}
    \caption{Ethio-ASR benchmark results (94M and 600M models, Amharic and Oromo) across Jetson Orin NX and Raspberry Pi 5, reporting WER\textsubscript{norm}, inference time per utterance, and peak RAM. Colour intensity encodes magnitude independently per metric.}
    \label{fig:asr_eval}
\end{figure*}
Inference latency is the dominant performance differentiator across tiers. On Tier~B, the 94\,M model achieves $\sim$47\,ms per utterance and the 600\,M achieves $\sim$478\,ms; on Tier~A, these increase to 3.05\,s and 15.2\,s respectively. The 94\,M model remains above real-time on Tier~A (approximately 3$\times$ headroom) and is suitable for near-real-time deployment. The 600\,M model at $\sim$15\,s per utterance on Tier~A operates below real-time and is appropriate only for offline batch transcription. Oromo exhibits substantially higher error rates across both tiers and model sizes (WER\textsubscript{norm} $\approx$0.705--0.763), a result attributable to limited training data availability rather than hardware or quantization effects.

\begin{table*}[!t]
\renewcommand{\arraystretch}{1.3}
\caption{Ethio-ASR Benchmark, Amharic \& Oromo, Both Tiers}
\label{tab:asr}
\centering
\begin{tabular}{lllrrrr}
\toprule
\textbf{Model} & \textbf{Lang} & \textbf{Device} & \textbf{WER\textsubscript{norm}} & \textbf{CER} & \textbf{Inf mean (s)} & \textbf{Power (W)} \\
\midrule
94M  & Amharic & Jetson (GPU) & 0.288 & 0.184 & 0.047 & 36.5 \\
94M  & Amharic & Pi 5 (CPU)   & 0.291 & 0.186 & 3.05  & 10.2 \\
600M & Amharic & Jetson (GPU) & 0.228 & 0.163 & 0.478 & 33.9 \\
600M & Amharic & Pi 5 (CPU)   & 0.252 & 0.166 & 15.23 & 10.6 \\
\midrule
94M  & Oromo   & Jetson (GPU) & 0.725 & 0.260 & 0.048 & 36.6 \\
94M  & Oromo   & Pi 5 (CPU)   & 0.725 & 0.260 & 2.92  & 10.3 \\
600M & Oromo   & Jetson (GPU) & 0.705 & 0.249 & 0.475 & 34.2 \\
600M & Oromo   & Pi 5 (CPU)   & 0.763 & 0.261 & 15.00 & 10.1 \\
\bottomrule
\end{tabular}
\end{table*}

\subsection{Deployment Recommendations}

Based on the evaluation results, the following deployment configurations
are recommended:

\begin{itemize}

  \item \textbf{Language model --- interactive voice agents:}
  \texttt{gemma-4-E2B-it} at Q4\_K\_M is the recommended configuration
  on both tiers, offering the highest decode throughput (28.8\,t/s on
  Tier~B, 7.5\,t/s on Tier~A), the highest topic classification accuracy
  (89.2\%), and a PPL drift of only 0.63\% relative to BF16. It is
  particularly strong on Nigerian Pidgin (96.2\%) and Igbo (87.5\%),
  making it the preferred choice for West African deployment contexts.

  \item \textbf{Language model --- memory-constrained deployments:}
  \texttt{gemma-2-2b-it} at Q4\_K\_M is the recommended lightweight
  alternative (1.59\,GiB, 3,515\,MiB RSS), with a topic accuracy of
  86.0\%. Its 1.59 GiB file and 3,515 MiB resident memory are the smallest of any evaluated configuration, making it the preferred choice where memory is the binding constraint. 

  \item \textbf{Language model --- highest fidelity to full-precision:}
  \texttt{gemma-4-E2B-it} at Q8\_0 is recommended where near-lossless
  quantization behaviour is required. It matches the BF16 reference
  almost exactly (89.0\% accuracy vs 88.8\%, PPL drift 0.44\%) while
  decoding 69\% faster than BF16 at roughly half the memory
  (7,565\,MiB vs 13,508\,MiB RSS).

  \item \textbf{Language model --- permissively licensed deployments:}
    \texttt{Qwen2.5-3B-Instruct} at Q4\_K\_M is recommended where a
    permissive license is required: it is the only evaluated model
    released under Apache~2.0. It achieves 82.5\% accuracy at a file
    size of 1.80\,GiB, with per-language PPL drift confined to a narrow
    $+1.35$ to $+4.20\%$ band, accepting a modest accuracy trade-off
    relative to \texttt{gemma-4-E2B-it}.
  
  \item \textbf{ASR:} The 94\,M Ethio-ASR model is recommended for
  real-time deployment on both tiers, achieving near-real-time
  performance on Tier~A (3.05\,s per utterance, approximately 3$\times$
  headroom) with a normalised WER of 0.291 on Amharic. The 600\,M model
  is recommended exclusively for offline batch transcription on Tier~A,
  where its marginal Amharic quality improvement (WER 0.252 vs 0.291)
  justifies the 15.23\,s per utterance inference cost.

  \item \textbf{Tier~A latency constraint:} The first-response latency
  of 1.6--2.2\,s on the Raspberry Pi~5 (a consequence of the
  24--30$\times$ prefill throughput degradation relative to Tier~B)
  constitutes the primary user-experience constraint for interactive
  deployment. Prompt-length reduction, retrieval context constraints,
  and prompt caching are the primary mitigation strategies. For
  applications where sub-second response initiation is required,
  Tier~B deployment is recommended.

\end{itemize}

\section{Conclusion}
\label{sec:conclusion}

The Offline AI Modules workstream delivers a credible, well-documented
foundation for voice-first offline AI on low-cost African-market hardware.
By combining a modular reference architecture, a reproducible hardware
BOM, a quantization pipeline, and a multilingual benchmarking methodology,
the workstream creates the conditions for community-wide progress rather
than isolated proof-of-concepts.

The benchmark results establish the first quantitative baseline for this
stack. \texttt{gemma-4-E2B-it} at Q4\_K\_M achieves 28.8\,t/s on Tier~B
with 89.2\% topic classification accuracy across five languages and a PPL
drift of 0.63\%, while all three models run within budget on Tier~A at
6.1--7.5\,t/s. The 94\,M Ethio-ASR model confirms that the full
voice-first pipeline is feasible on affordable hardware without internet
connectivity. First-response latency on Tier~A (1.6--2.2\,s) and Oromo
ASR quality (WER 0.705--0.763) remain the primary open challenges,
pointing respectively to prompt optimisation and African-language data
availability as the next priorities.

\appendices
\section{Quantization Format Quick-Reference}\label{app:quant}

Q4\_K\_M is the recommended default for Tier~A community deployments
  ($\sim$1.9\,GB for a 3B model; measured task accuracy matched or
  exceeded the full-precision reference for all three evaluated models,
  with macro PPL drift between 0.6\% and 6.4\%). Q5\_K\_M offers near-FP16 quality at modest size cost ($\sim$2.2\,GB). Q8\_0 is effectively lossless and targets Tier~B (Jetson Orin NX). F16 is the calibration baseline and is not suitable for deployment on Tier~A hardware.

\section{Benchmark JSON Schema Summary}\label{app:schema}

Each run produces a run-header record (device, pipeline, model, timestamp) and one per-interaction metrics record (end-to-end latency, ASR latency, LM first-token latency, LM throughput, TTS latency, peak RAM, errors). The CSV schema is a flattened view of the same fields, one row per interaction.

\section{Per-Class Precision, Recall, and F1}
\label{app:class_f1}

Table~\ref{tab:class_f1} reports precision, recall, and F1 per topic
class for all evaluated model and quantization combinations on the
MasakhaNEWS benchmark. Sports achieves the highest F1 across all
configurations (91.1--96.0\%), while health and politics are
consistently harder. Politics recall is uniformly 95.0--100.0\% across
all models and formats, indicating a systematic tendency to over-predict
the politics class, which correspondingly reduces its precision
(64.1--79.5\%). This pattern is stable across quantization levels and
does not appear to be induced by compression.

\begin{table*}[!t]
\renewcommand{\arraystretch}{1.2}
\caption{Per-Class Precision, Recall, and F1 (\%) --- Jetson Orin NX}
\label{tab:class_f1}
\centering
\begin{tabular}{llllrrr}
\toprule
\textbf{Model} & \textbf{Quant} & \textbf{Class} & \textbf{Support} & \textbf{Precision (\%)} & \textbf{Recall (\%)} & \textbf{F1 (\%)} \\
\midrule
Gemma-4-E2B-it & BF16    & entertainment & 100 & 96.3 & 79.0 & 86.8 \\
               &         & health        & 100 & 94.2 & 81.0 & 87.1 \\
               &         & politics      & 100 & 75.2 & 100.0 & 85.8 \\
               &         & sports        & 100 & 96.9 & 95.0 & 96.0 \\
\midrule
Gemma-4-E2B-it & Q8\_0   & entertainment & 100 & 96.3 & 79.0 & 86.8 \\
               &         & health        & 100 & 94.3 & 82.0 & 87.7 \\
               &         & politics      & 100 & 75.8 & 100.0 & 86.2 \\
               &         & sports        & 100 & 96.9 & 95.0 & 96.0 \\
\midrule
Gemma-4-E2B-it & Q5\_K\_M & entertainment & 100 & 90.6 & 77.0 & 83.2 \\
               &          & health        & 100 & 94.3 & 83.0 & 88.3 \\
               &          & politics      & 100 & 77.3 & 99.0 & 86.8 \\
               &          & sports        & 100 & 95.9 & 93.0 & 94.4 \\
\midrule
Gemma-4-E2B-it & Q4\_K\_M & entertainment & 100 & 90.2 & 83.0 & 86.5 \\
               &          & health        & 100 & 93.3 & 83.0 & 87.8 \\
               &          & politics      & 100 & 79.5 & 97.0 & 87.4 \\
               &          & sports        & 100 & 96.9 & 94.0 & 95.4 \\
\midrule
Gemma-2-2b-it  & F16     & entertainment & 100 & 95.7 & 66.0 & 78.1 \\
               &         & health        & 100 & 92.5 & 74.0 & 82.2 \\
               &         & politics      & 100 & 65.4 & 100.0 & 79.1 \\
               &         & sports        & 100 & 92.6 & 87.0 & 89.7 \\
\midrule
Gemma-2-2b-it  & Q8\_0   & entertainment & 100 & 95.7 & 66.0 & 78.1 \\
               &         & health        & 100 & 92.3 & 72.0 & 80.9 \\
               &         & politics      & 100 & 64.5 & 100.0 & 78.4 \\
               &         & sports        & 100 & 92.6 & 87.0 & 89.7 \\
\midrule
Gemma-2-2b-it  & Q5\_K\_M & entertainment & 100 & 94.5 & 69.0 & 79.8 \\
               &          & health        & 100 & 93.5 & 72.0 & 81.4 \\
               &          & politics      & 100 & 64.9 & 100.0 & 78.7 \\
               &          & sports        & 100 & 95.6 & 87.0 & 91.1 \\
\midrule
Gemma-2-2b-it  & Q4\_K\_M & entertainment & 100 & 91.0 & 81.0 & 85.7 \\
               &          & health        & 100 & 93.8 & 76.0 & 84.0 \\
               &          & politics      & 100 & 73.6 & 95.0 & 83.0 \\
               &          & sports        & 100 & 93.9 & 92.0 & 92.9 \\
\midrule
Qwen2.5-3B     & F16     & entertainment & 100 & 94.9 & 75.0 & 83.8 \\
               &         & health        & 100 & 95.6 & 65.0 & 77.4 \\
               &         & politics      & 100 & 64.9 & 100.0 & 78.7 \\
               &         & sports        & 100 & 97.9 & 94.0 & 95.9 \\
\midrule
Qwen2.5-3B     & Q8\_0   & entertainment & 100 & 94.9 & 75.0 & 83.8 \\
               &         & health        & 100 & 95.7 & 67.0 & 78.8 \\
               &         & politics      & 100 & 66.2 & 100.0 & 79.7 \\
               &         & sports        & 100 & 97.9 & 94.0 & 95.9 \\
\midrule
Qwen2.5-3B     & Q5\_K\_M & entertainment & 100 & 94.8 & 73.0 & 82.5 \\
               &          & health        & 100 & 93.8 & 61.0 & 73.9 \\
               &          & politics      & 100 & 64.1 & 100.0 & 78.1 \\
               &          & sports        & 100 & 96.9 & 95.0 & 96.0 \\
\midrule
Qwen2.5-3B     & Q4\_K\_M & entertainment & 100 & 97.0 & 65.0 & 77.8 \\
               &          & health        & 100 & 90.0 & 72.0 & 80.0 \\
               &          & politics      & 100 & 64.9 & 100.0 & 78.7 \\
               &          & sports        & 100 & 96.9 & 93.0 & 94.9 \\
\bottomrule
\end{tabular}
\begin{minipage}{\textwidth}
\vspace{4pt}
\footnotesize\textit{Support is 100 items per class across all five languages. Politics
recall is uniformly high (95.0--100.0\%) across all configurations, with
correspondingly reduced precision (64.1--79.5\%), indicating a systematic
over-prediction tendency that is stable across quantization levels.}
\end{minipage}
\end{table*}

\end{document}